\documentclass[runningheads]{llncs}

\usepackage{eccv}

\usepackage{eccvabbrv}

\usepackage{graphicx}
\usepackage{booktabs}

\usepackage[accsupp]{axessibility}  

\usepackage{amsmath}
\usepackage{amssymb}
\usepackage{amsfonts}
\usepackage{bm}
\usepackage{comment}
\usepackage{color}
\usepackage{array}
\usepackage{colortab}
\usepackage{colortbl}
\usepackage{arydshln}
\usepackage{multirow}
\usepackage{wrapfig}
\newcommand{\ryoshiha}[1]{\textcolor{black}{#1}} 
\newcommand{\ryoshihab}[1]{\textcolor{black}{#1}} 
\newcommand{\kurdyla}[1]{\textcolor{black}{#1}} 
\newcommand{\ryoshihac}[1]{\textcolor{black}{#1}} 
\newcommand{\hk}[1]{\textcolor{black}{#1}} 
\newcommand{\bhline}{\noalign{\hrule height 1.25pt}} 
\newcommand{\mysubsection}[1]{\vspace{-2mm}\subsection{#1}\vspace{-0mm}}
\newcommand{\mysection}[1]{\vspace{-3mm}\section{#1}\vspace{-2mm}}
\newcommand{\gr}{\color[rgb]{0.6,0.6,0.6}} 

\begin{document}

\title{Exploring Limits of Diffusion-Synthetic Training \\ with Weakly Supervised Semantic Segmentation} 

\titlerunning{Exploring Limits of Diffusion-Synthetic Training with WSSS}

\author{Ryota Yoshihashi \and
Yuya Otsuka \and
Kenji Doi \and 
Tomohiro Tanaka \and
Hirokatsu Kataoka }

\authorrunning{R. Yoshihashi et al.}

\institute{LY Corporation}

\maketitle

\begin{abstract}
The advance of generative models for images has inspired various 
training techniques for image recognition utilizing synthetic images.
In semantic segmentation, one promising approach is extracting pseudo-masks from attention maps in text-to-image diffusion models, which enables real-image-and-annotation-free training.
However, the \ryoshihab{pioneering} training methods using the diffusion-synthetic images and pseudo-masks, e.g., DiffuMask have limitations in terms of mask quality,
scalability, and ranges of applicable domains.
\ryoshihac{To address these limitations, we propose a new framework to view diffusion-synthetic semantic segmentation training as a weakly supervised learning problem, where we utilize potentially inaccurate attentive information within the generative model as supervision.
Motivated by this perspective, we first introduce reliability-aware robust training, \ryoshihab{originally used as a classifier-based WSSS method}, with modification to handle generative attentions. 
Additionally, we propose techniques to boost the weakly supervised synthetic training: we introduce prompt augmentation by synonym-and-hyponym replacement, which is data augmentation to the prompt text set to scale up and diversify training images with limited text resources.}
Finally, LoRA-based adaptation of Stable Diffusion enables the transfer to a distant domain, e.g., auto-driving images. 
Experiments in PASCAL VOC, ImageNet-S, and Cityscapes show \kurdyla{that} our method effectively closes gap between real and synthetic training in semantic segmentation. 
  \vspace{-2mm}
\end{abstract}

\vspace{-5mm} \mysection{Introduction}
\ryoshiha{Advanced generative models have enabled high-fidelity and controllable synthesis of images.
Diffusion models \cite{ho2020denoising} can generate high-quality images without adversarial training \cite{goodfellow2014generative}, which often makes the optimization tricky.
Large-scale image-caption pair datasets \cite{schuhmann2021laion} enabled text-to-image (text2img) generation \cite{ramesh2021zero,nichol2022glide,rombach2022high}, which is useful to control generated image contents by \kurdyla{using} free-form textual prompts.
These generative trends inspired research in discriminative learning to train models only using generated images without relying on manually collected and annotated image datasets \cite{tian2023stablerep,sariyildiz2023fake}.}

\ryoshiha{In semantic segmentation, it is essential to generate pixel-wise semantic labels in addition to images for enabling synthetic training. Fortunately, text2img diffusion models naturally can extract pixel-semantic labels by utilizing their internal text-image cross-attention layers \cite{wu2023diffumask}. In the generation processes, the cross-attention layers are used to relate the text tokens to the elements of latent image representations to reflect the prompt. In other words, the cross-attention layers solve the grounding between words and pixels in generated images \cite{tang2022daam}, and they can serve as sources of localization information \cite{hertz2022prompt,tian2023diffuse,zhou2022maskclip}. These text-image cross-attention maps, \ryoshihac{when paired with generated images}, can be used as supervisory signals for segmentation.
A pioneering work in this direction is DiffuMask \cite{wu2023diffumask}, which generates training images and labels for each class of the target benchmark (e.g., PASCAL VOC \cite{everingham2015pascal}) with Stable Diffusion (SD) \cite{rombach2022high}.}

\ryoshiha{However, DiffuMask \cite{wu2023diffumask} has several limitations in synthetically training segmentation models. 
First, the \kurdyla{quality of synthetic labels}, which is not always as accurate as human-annotated ones, affects the segmentation models' performances directly.
Second, the image synthesis largely relies on template-based artificial prompts only related to one among the target classes (e.g., 20 classes in PASCAL VOC), which limits the scale and diversity of training data.
Third, applying the segmentation models to domains distant from the original SD's training domain, i.e., web-crawled images, \kurdyla{is} hard.
DiffuMask was only applicable to small subclasses of all defined classes when it was transferred to Cityscapes \cite{cordts2016cityscapes}, a dataset for urban auto-driving.}

\ryoshiha{We hypothesize that the first and largest part of these limitations comes from fully supervised segmentation training \cite{long2015fully,deeplabv3plus2018,cheng2022masked} using the synthetic data that are not yet perfect.
\kurdyla{In contrast}, we frame the synthetic training as a weakly supervised task where we need to train accurate segmentation from possibly inaccurate generated labels. This framing is reasonable because text2img generators, including SD, can be seen as text-supervised models, and synthetic training using them can be regarded as text-supervised weakly supervised learning for segmentation.
Weakly supervised semantic segmentation (WSSS) with real images \kurdyla{has been well studied \cite{oquab2015object,zhou2016learning,rong2023boundary}}, and especially, we incorporate the idea of robust co-training for WSSS \cite{rong2023boundary} with modifications for synthetic training. \ryoshihac{One notable difference is that conventional WSSS mainly utilized the classifiers' activation maps (CAMs), but we replace them with generative attentions, whose statistics are different as discussed later. However, we successfully overcome the difference by tailoring an adaptive threshold strategy for the SD's generative attentions.}}

\ryoshiha{We further explore the limitations \kurdyla{from} the two perspectives: scalability and range of application.
To easily scale up the size of the synthetic training data,
we propose prompt augmentation by synonym-and-hyponym replacement, which is a data augmentation technique in the prompt space rather than in the image space and is useful to inflate the limited text corpus to seed more generated images. 
 To widen the range of applications, we further incorporate an adaptation technique with LoRA \cite{hu2021lora}, which \kurdyla{quickly} finetunes SD with a small amount of target-domain real unannotated images. 
 Here, our finding is that SD does not lose its mask-producing ability after LoRA-based finetuning even though the target-domain images are not paired with textual prompts, which successfully perform unsupervised domain adaptation.}

Incorporating these ideas, we developed our diffusion-synthetically trained semantic segmentation method {\it Attn2mask}.
 In experiments, we verify Attn2mask's effectiveness by achieving 62.2 mIoU in the PASCAL VOC segmentation task, which is accurate for not using any real images or human annotation.
Attn2mask is also applicable to a larger-scale scenario in ImageNet-S 1k-class segmentation,
where it performs sufficiently close to the semi-supervised BigDatasetGAN \cite{li2022bigdatasetgan} generator.
Furthermore, the adaptation of SD is shown to be beneficial for distant-domain transfer for the Cityscapes driving images, although it \kurdyla{slightly differs} from the spirit of real-image-free training.

\hk{Our contributions are summarized as follows:} 

\noindent\hk{\textbf{{Conceptual contribution.}} We connect the ideas of diffusion-synthetic training and WSSS for the first time. In other words, we demonstrate that even attentive masks from the strong SD model can be regarded as weak supervision and have room for improvement with WSSS training techniques.}

\noindent\hk{\textbf{{Technical contribution.}} We propose a bag of approaches to push the limits of diffusion-synthetic WSSS, namely, adaptive threshold for generative attention maps, prompt augmentation to boost the diversity of synthetic datasets, and domain-adaptive
LoRA to make SD applicable to distant-domain tasks. The effects of each of them are validated by the experiments.}

\noindent\hk{\textbf{{Experimental contribution.}} We demonstrate Attn2mask's effectiveness \\equipped with these ideas in the experiments, especially by achieving 62.2 mIoU with ResNet50-based networks, which outperforms the existing state of the arts in synthetically trained semantic segmentation without human-annotated labels.}

\vspace{-1mm}\mysection{Related work}
\vspace{-1mm}\subsubsection{Segmentation with diffusion models}
Several studies have tried to leverage the diffusion models' excellent
image-generation performance in segmentation.
There are two lines of research: reusing diffusion U-Net for segmentation tasks and generating training image-label pairs for segmentation. 
For the former line, denoising diffusion probabilistic models (DDPMs) can be used as pre-trained feature extractors for semantic segmentation \cite{baranchuk2022label}.
ODISE \cite{xu2023open} and VPD \cite{zhao2023unleashing} similarly fine-tune
a text2img diffusion model for segmentation, but it enables phrase-based segmentation by reusing a text encoder in the text2img model.
\ryoshihac{Similar approaches are seen in single-object localization \cite{zhao2023generative} or object discovery \cite{ma2023diffusionseg}.}
These usages of text2img models as visual-linguistic backbones can be seen as a generative counterpart of contrastive visual-linguistic foundation models, e.g., CLIP \cite{radford2021learning}, which also can be repurposed for segmentation tasks \cite{rao2022denseclip}.
However, all of the aforementioned methods rely on manually annotated pixel labels for training.
\ryoshihab{Very recently, diffusion-based training-free methods were proposed \cite{karazija2023diffusion,wang2023diffusion}. While they are attractive by omitting the additional training phase, their segmentation accuracies tend to be suboptimal; \ryoshihac{in contrast, we emphasize that robust re-training based on WSSS techniques is our source of improvement}.}

\ryoshiha{We call the latter line of research {\it diffusion-synthetic training}}.
For example, Grounded Diffusion \cite{li2023guiding} trains image-and-mask generation models by leveraging a pre-trained Mask R-CNN with pixel labels.
\ryoshihac{FreeMask \cite{yang2024freemask} augmented real image-and-mask datasets using mask-to-image generation.}
We refer to \kurdyla{these kinds} of approaches {\it semi-supervised} mask generation \kurdyla{because of} their dependency on annotated training images on at least partial classes.
DiffuMask \cite{wu2023diffumask} is closer to ours in the sense of {\it weakly supervised} mask generation, where masks are automatically acquired via text2img generation training, but it only supports single-object image generation.
Dataset Diffusion \cite{nguyen2023dataset} is another work in this line. It enables multi-class image-mask generation using real image captions.
In the concept level, ours differs in that we frame the problem in WSSS to focus on improving possibly inaccurate generated labels in the segmentation training phase, while the two studies adopted fully supervised segmentation in the downstream training. In more detail, 1) we use prompt augmentation to acquire more diversified and contextualized prompts instead of template-based or real-caption-based prompts, and 2) we omit iterative self-training \cite{nguyen2023dataset} or cross-validation-based synthesized image curation \cite{wu2023diffumask} thanks to the co-training that is robust even with single-step training.

\vspace{-5mm}\subsubsection{Training with generated images}
The idea of exploiting free annotations with synthesized images dates back \kurdyla{to} before the large-scale generative models:
for example, computer-graphics-based dataset creation using game engines was examined \cite{richter2016playing,ros2016synthia}.
\ryoshihac{Cut-paste-based learning \cite{dwibedi2017cut,ge2023beyond} can be seen as a simpler form of the dataset-generation methods}.
More primitively, even segmentation of mathematically drawn contours can help real segmentation as a pretraining task \cite{kataoka2022replacing,shinoda2023segrcdb}. 
However, such rendered images tended to have gaps from real ones, which needed supervised retraining or unsupervised adaptation techniques \cite{hoffman2018cycada} to utilize them fully to achieve preferable real-environment performances.
In contrast, we exploit modern generative models' high-fidelity generation ability that narrows the gap without any adaptation \kurdyla{at least in} generic settings such as PASCAL VOC. 
Prior to the diffusion models, GAN-based image-and-mask generation was also tried \cite{li2022bigdatasetgan} but manual annotation \kurdyla{was still required} for partial classes to fine-tune the GANs for mask generation.
In addition to segmentation, classification \cite{sariyildiz2023fake} and unsupervised representation learning \cite{tian2023stablerep,zhang2023free} with generated images have been studied, and they report
 performances gradually approaching those of real-dataset-based counterparts, which is motivating for studies in synthetic training including ours.

\vspace{-5mm}\subsubsection{Weakly supervised semantic segmentation}
Mainstream WSSS approaches exploit image-level class labels.
Two-stage training that first trains an image-level classifier and then re-trains another segmentation model using the classifier's class-activation maps (CAM) \cite{zhou2016learning} as weak labels is the most straightforward but still well-used strategy.
Among the two-stage methods, existing studies attempted techniques for better CAM generation \cite{wang2020self,jo2021puzzle,li2021pseudo,chen2022class} and robust segmentation training against weak labels \cite{liu2022adaptive,li2022uncertainty,rong2023boundary}.
From the viewpoint of WSSS, we replace classifier-based CAMs with generative attentions combined with generated images. In addition, we incorporate the latest robust segmentation-training method \cite{rong2023boundary}, which we \kurdyla{find} to be beneficial in our training scenario.
\ryoshihac{A preprint concurrent with ours leveraged ControlNet \cite{zhang2023adding} for data augmentation in WSSS \cite{wu2023image}, but this approach is largely different from ours in terms of not supporting pure synthetic training without real images.} 
\ryoshihac{Recent methods integrated the Segment-Anything Model (SAM) \cite{kirillov2023segment} in WSSS \cite{sun2023alternative,yang2024foundation}.
SAM in itself is a {strongly} supervised for segmentation because its pretraining involves human pixel-annotation.
While ours similarly relies on SD, it is { weakly} supervised for segmentation since it depends on image-text pairs but no pixel-level intervention; one can be less reluctant to call our method weakly supervised even considering the pre-training.}

\vspace{-3mm}\subsubsection{Unsupervised segmentation and text-based segmentation}
Recently, segmentation studies diverged in various training strategies.
For example, unsupervised semantic segmentation does not use any form of supervision in the downstream segmentation training \cite{yu2021unsupervised,cho2021picie}.
Our Attn2mask may be seen as unsupervised in the sense that it is annotation-free in the downstream domains, but it operates under the stricter constraint of unavailability of downstream-domain (real) images than usual unsupervised methods.
Text-based segmentation exploits vision-language pretraining \cite{rao2022denseclip,zhou2022maskclip,shin2022reco}, which often transfers well in the downstream segmentation tasks without fine-tuning. 
Attn2mask is related to these methods in the usage of vision-language pretraining except for the difference between generative (i.e., SD) and discriminative (i.e., CLIP) pretraining.

\begin{figure}[t]
    \centering
    \includegraphics[width=1\linewidth]{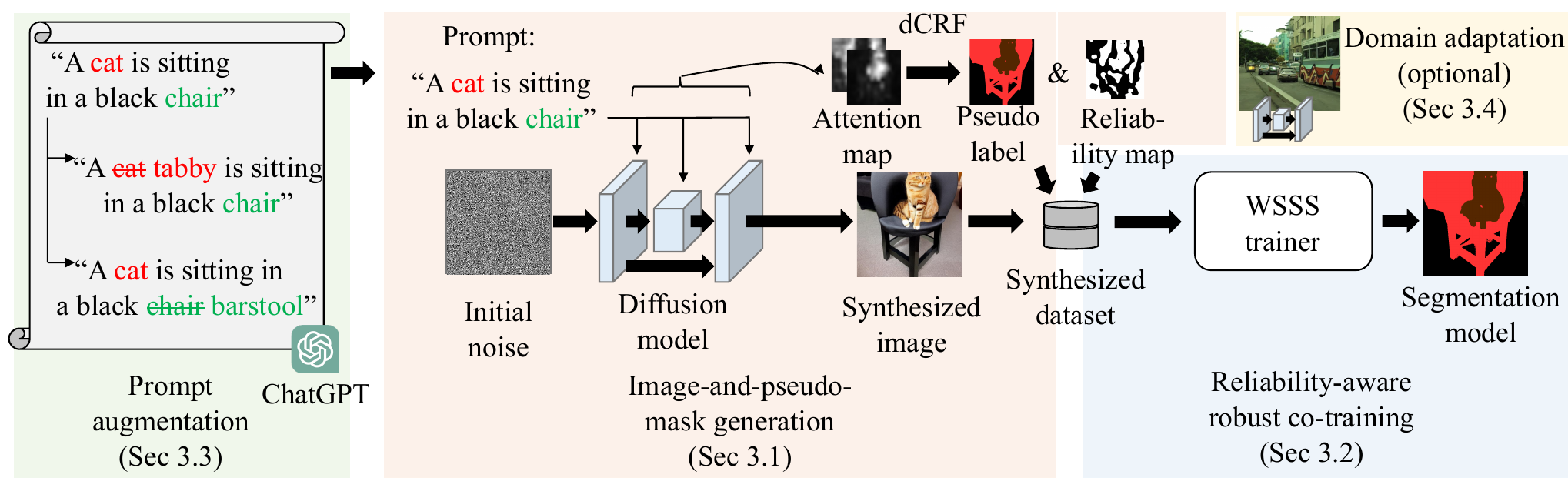}
    \vspace{-6mm}
    \caption{Overview of the proposed method {\it Attn2mask}.
    The major parts are image-and-pseudo-mask generation and reliability-aware robust co-training. Additionally, we adopt prompt augmentation, a data-augmentation technique for prompts. Domain adaptation is optionally supported for far-domain transfer.}
    \label{fig:overview}\vspace{-6mm}
\end{figure}
\mysection{Proposed method Attn2mask}
\ryoshiha{An overview of Attn2mask is shown in Fig.\ref{fig:overview}.
Its major parts are image-and-pseudo-mask generation and segmentation training. In the segmentation-training, we utilize a robust co-training strategy to handle possibly inaccurate generated pseudo-masks. Additionally, we adopt prompt augmentation, a data-augmentation technique for prompts to generate images rather than to the generated images.}

\vspace{-3mm}\mysubsection{Image-and-pseudo-mask generation}\vspace{-2mm}
To extract semantic regions from diffusion-synthesized images, we need to compute the contributions of each word in the prompts on each location in images.
Fortunately, image-text cross-attention, \kurdyla{commonly} used in most diffusion models including SD, is naturally available for this purpose.  
Transformer-style attention modules can be written as follows:
\begin{eqnarray}
 \text{Transformer}(Q, K, V) &=& A(Q, K) V, \\
 A(Q, K) &&= \text{Softmax}_{WH}(\frac{QK^T}{\sqrt{\text{dim}(K)}}),
\end{eqnarray}
where Q, K, and V are the query, key, and value, each of which is a sequence of vectors.
In using this as text-to-image cross-attention,
Q is computed from the image embedding $\bm{f} \in \mathbb{R}^{W \times H \times C}$ whose size is $W \times H$ and dimensionality is $C$, and K and V are computed from the text embedding $\bm{s} \in \mathbb{R}^{L \times D}$ whose length is $L$ and dimensionality is $D$.
This makes the attention maps' dimensionality
\begin{eqnarray}
    A(Q(\bm{s}), K(\bm{f})) &\in& \mathbb{R}^{W \times H \times L},
\end{eqnarray}
where the element at $(x, y, t)$ can be interpreted as the amount of contribution from the $t$-th text token to the location in image coordinate $(x, y)$.

From the attention maps $A$, we extract channels that are relevant to the classes of interest to use them as supervisory masks.
Given a token sequence of the input prompt,
we denote the positions of the relevant tokens of class $c$ by
$\tau_c = [t_{c1}, t_{c2}, ..., t_{cn_c}]$,
allowing multiple occurrences of the relevant tokens for a single class.
\ryoshihac{The ``relevance'' of words and classes can be set manually by any rule-based matching, and in our implementation, we list singular and plural forms of synonyms for a class's display name and extracted exact matches.}
\ryoshihab{This relevance-based attention extraction can be denoted as}
\begin{eqnarray}\label{eq:aggre}
    A_c = \frac{1}{n_c}\sum_{t \in \tau_c} {A(Q(\bm{s}), K(\bm{f}))_t} &\in& \mathbb{R}^{W \times H},
\end{eqnarray}
where we denote the operation to extract $t$-th channels of the attention map by $A(\cdot, \cdot)_t$.

We add one more probability map for the background as a class defined by
\begin{eqnarray}
A_0(x, y) &=& 1.0 - \max_{c \in [1, N]}A_c(x, y) - \beta,
\end{eqnarray}
which represents the absence of activations for any other classes.
Here, $N$ is the number of (non-background) classes, and $\beta$ is a hyperparameter to control the strength of the background prior; larger $\beta$ brings smaller portions of the background area in labeling results.
The motivation behind this special treatment for the backgrounds is that they tend to be generated implicitly without concrete indication in the prompts, which makes class-word-based aggregation in Eq. \ref{eq:aggre} difficult.

After attention maps $A = [A_0, A_1, ..., A_N]$ are gathered, we apply binarization to them to acquire multi-class discrete pseudo-labels.
We use the densely connected conditional random field (dCRF) \cite{krahenbuhl2011efficient}, a non-learning-based labeling algorithm. This is also useful to improve mask quality by considering the color similarity and local smoothness in labeling.
\ryoshihab{Although the prior work \cite{wu2023diffumask} applied dCRF after threshold-based binarization of FG/BG masks,
we use dCRF in multi-class, continuous-probability-based labeling that is the original usage in \cite{krahenbuhl2011efficient}.}
These generation processes are iterated for a given set of prompts, and generated images and masks are stored in the disk as a dataset.

\begin{figure}[t]
    \hspace{0mm}\begin{minipage}{.52\hsize}
       \centering
        \includegraphics[width=1.\linewidth]{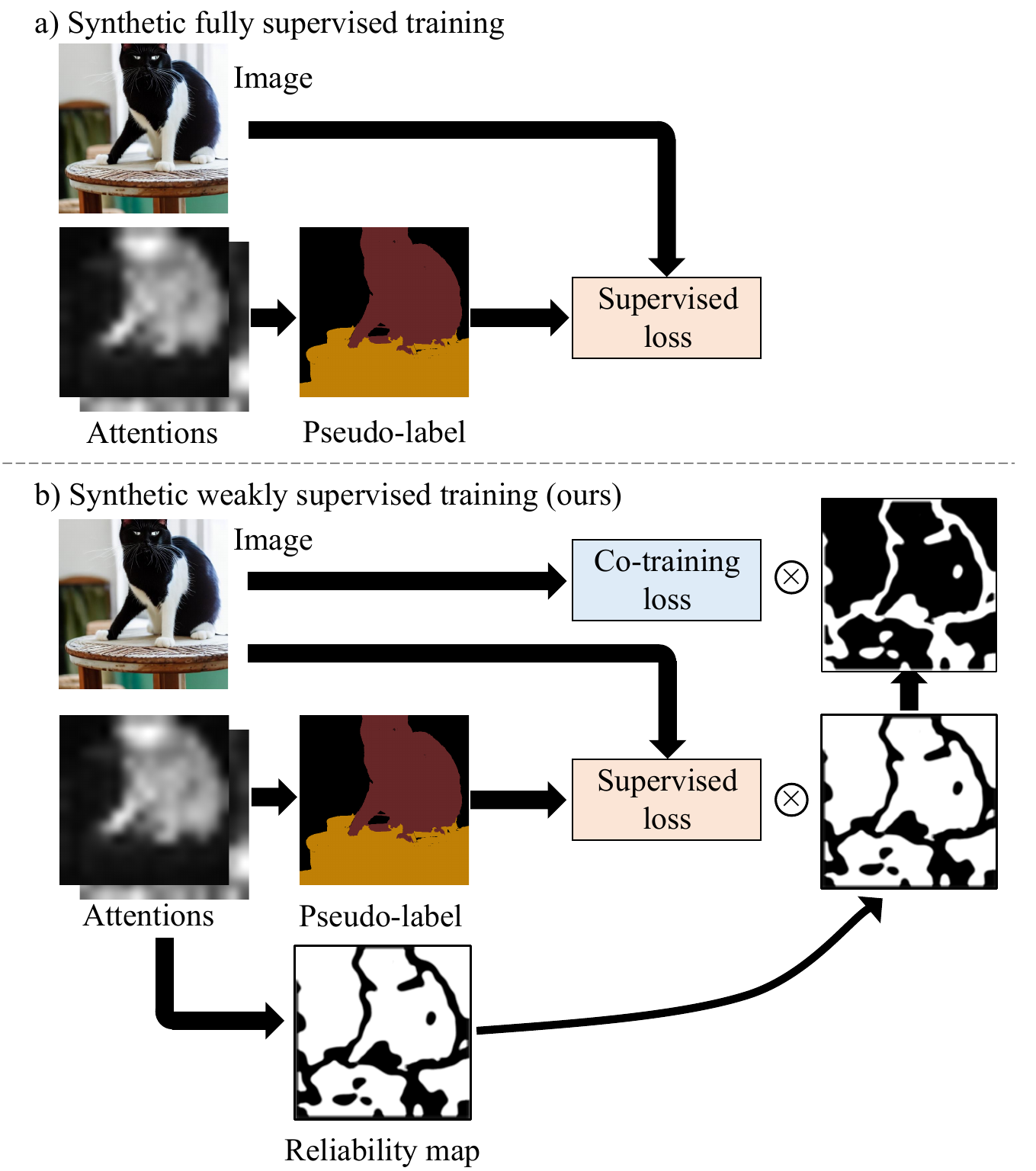}
        \vspace{-4mm}
        \caption{Fully supervised and weakly supervised training schemes with synthetic training pseudo labels.}
        \label{fig:beco}
        \end{minipage}
    \hfill
    \hspace{0mm}\begin{minipage}{.46\hsize}
        \centering
    \includegraphics[width=.90\linewidth]{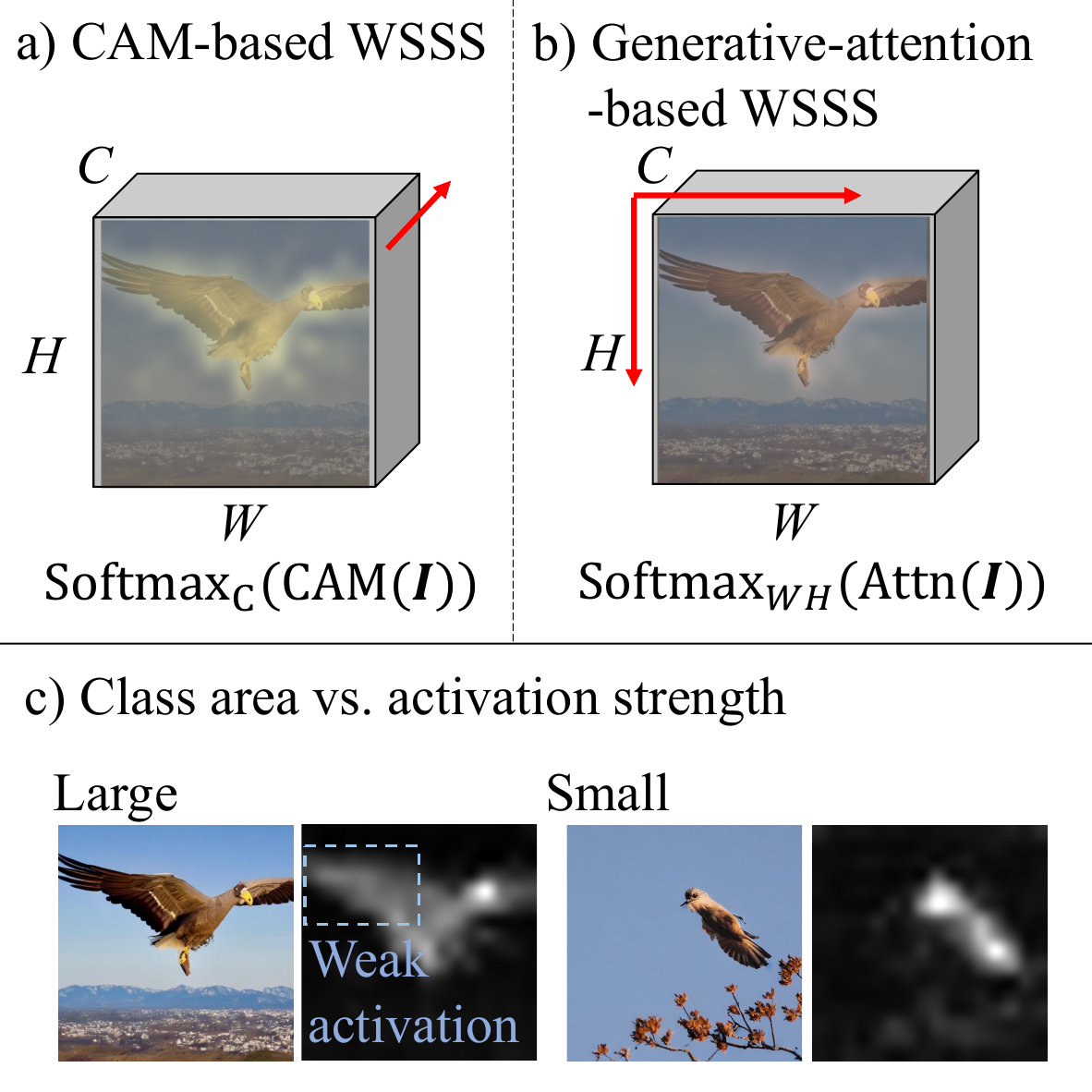}
    \vspace{-1mm}
    \caption{\ryoshihac{A difference between a) CAM and b) generative attentions used as pseudo-masks lies in the direction of Softmax operation. 
    The spatial Softmax in the generative attentions is prone to cause weak activations simply by the large area that a class  (in the examples {\it bird}) occupies. 
    We mitigate this problem with adaptive thresholding.}}
    \label{fig:cam}
    \end{minipage}\vspace{-6mm}
\end{figure}

\vspace{-3mm}\mysubsection{Reliability-aware robust training}\label{sec:cot}\vspace{-3mm}
Once our images and pseudo-labels are generated, any supervised segmentation models are generally trainable with them. 
However, naive training of supervised learners may suffer since the attention-based pseudo-labels are not always perfectly accurate;
robust learners that are tolerant of label noises would help.
Thus, we adopt a robust training algorithm
from BECO \cite{rong2023boundary}, a recent \ryoshihab{WSSS} method \ryoshiha{with modification to utilize generative attention maps} as shown in Fig. \ref{fig:beco}.

BECO is an adaptive co-training method exploiting reliability maps
of the pseudo-labels: it performs supervised learning with full trust in the pseudo-labels within regions where the pseudo-labels are confident and performs co-training-based consistency regularization in regions where the pseudo-labels are unreliable.
In the original work, BECO was trained with reliability maps generated by processing the classifier's confidence score maps.
\ryoshihab{Rather than additionally train a classifier,} \ryoshihac{a naive approach might be binarizing } the attention maps with a constant threshold to use as the reliability map as follows:
\begin{eqnarray}
R(x, y) &=& \left\{
\begin{array}{ll}
1 & \text{if } \max_{c \in [0, N]}A_c(x, y) \geq r \\
0 & \text{otherwise},
\end{array}
\right.
\end{eqnarray}
where $r$ is a hyperparameter for the reliability threshold.

\ryoshihac{However, there is a difference in sources of pseudo-masks that poses a challenge to application of the reliability-based co-training: conventional WSSS has utilized CAM, which is computed by Softmax along the class axis as shown in Fig. \ref{fig:cam} a. The CAM is discriminatively optimized and therefore it is intuitive that we can interpret their raw values as reliability.
Generative attentions in SD differ in that their Softmax is along the spatial axes, which aims to allocate contributions from the text embedding based on the spatial constant-sum constraint, as shown in Fig. \ref{fig:cam} b.
A natural result of this is that a larger object is harder to cover with high activation values simply because the values are distributed across a broader region as seen in in Fig. \ref{fig:cam} c, which may be harmful by discarding a large portion of properly generated and pseudo-masked regions with a naive thresholding.}

\ryoshihac{Thus, we newly developed an adaptive threshold strategy that considers the attention strength, spatial distributions of the attention, and smoothed label assignments by dCRF in a joint manner as follows:}
\begin{eqnarray}
S &=& \text{dCRF}(A), \\
r(A, c) &=& \alpha \frac{\sum_{(x, y) \in \{(x, y) | S(x, y) = c\}}{A_{c}(x, y)}}{\sum_{(x, y) \in \{(x, y) | S(x, y) = c\}}{1}},
\end{eqnarray}
\ryoshihac{where $S \in \{0, 1, 2, ..., N\}^{W \times H}$ denotes labels assigned by dCRF, 
$r(A, c) \in [0, 1]^{W \times H}$ denotes the reliability-threshold maps, and $c \in {0, 1, 2, ..., N}$ denotes the class of interest. A hyperparameter $\alpha$ is introduced to control the overall hardness of the thresholds.
This threshold becomes smaller when the area of the class of interest used in the denominator $\sum_{(x, y) \in \{(x, y) | S(x, y) = c\}}{1}$ is larger, and is suitable to mitigate discarding the weak attention activations when the class areas are large.
These adaptive threshold maps are applied to the attention pseudo-masks with w.r.t. dCRF's selection of class labels in each location as }
\begin{eqnarray}
R(x, y) &=& \left\{
\begin{array}{ll}
1 & \quad \text{if } A_{S(x, y)}(x, y) \geq r(A, S(x, y)) \\
0 & \quad \text{otherwise},
\end{array}
\right.
\end{eqnarray}
and in this way we acquired the reliability map to switch on/off the co-training.

\vspace{-3mm}\mysubsection{Prompt augmentation}\label{sec:pa}\vspace{-2mm}
\ryoshiha{The abovementioned processes rely on a set of prompts to generate many diversified training samples. However, preparing a number of proper texts may be additional labor even though it is less hard than pixel annotation. Existing work utilized template-based prompt generation \cite{wu2023diffumask} or real captions from a captioning dataset \cite{nguyen2023dataset}, but they limited the total numbers of available training images to less than 100k.}

\ryoshiha{We take a simple but effective strategy for gathering prompt texts; we first curate real captions related to the target classes and next automatically augment them with {\it synonym-and-hyponym replacement}. Synonym replacement \cite{wei2019eda} is a technique for text data augmentation to replace words randomly with other words that have the same or similar meanings, for example ``bicycle'' by ``bike''. We additionally use the hyponyms s, for example ``owl'' for ``bird'', for replacement because they are useful to diversify generated images. Lists of synonyms and hyponyms for the 20 VOC classes are easily collected by asking ChatGPT \cite{chatgpt} ``\texttt{please raise fifteen examples of synonyms or hyponyms for  \$\{class\_names\}}''. \ryoshihab{We utilize the prompt augmentation in combination with image-space data augmentation.}}

\ryoshihac{The advantage of our prompt augmentation strategy is that it scales up a text corpus in a combinatorial manner:
Given $T$ texts and $C$ synonyms for a class, our prompt augmentation can generate $TC$ variations of prompts. 
Reviewing existing prompt strategies, Dataset Diffusion \cite{nguyen2023dataset} relied on the raw texts with the size $T$, and DiffuMask \cite{wu2023diffumask} added simple templates like ``a photo of [synonym],'' which makes the corpus scale $T + C$; both were bottlenecked with the corpus sizes.}

\vspace{-3mm}\mysubsection{Domain adaptive image-mask generation}\label{sec:da}\vspace{-2mm}
While image-mask generation with off-the-shelf SD generally performs satisfactorily, 
in out-of-domain applications, it may suffer from the domain gap between the generative model and the downstream tasks.
We notice that this is the case, for example, in applying Attn2mask to driving images in Cityscapes \cite{cordts2016cityscapes}.

To make Attn2mask applicable to new domains, we develop an adaptation method using unlabeled training images.
Here, we repurpose the instantiated generation technique DreamBooth \cite{ruiz2023dreambooth} for domain adaptation.
For a brief review, DreamBooth learns an instance token (often denoted by [V]) from a relatively small set of instance images. After fine-tuning with the instance-image set, [V] is usable as a new word in prompts to indicate the instance.
Using DreamBooth, we instead learn a domain token; by
training [V] not for a specific instance but for general training images in a new domain, SD can adapt to the domain, while \kurdyla{maintaining} the ability to reflect other prompt words to the image contents.

In implementation, we especially use DreamBooth-LoRA \cite{dblora}, which trains a low-rank adapter \cite{hu2021lora} that is a small sub-network instead of fine-tuning the whole network. This enables fast and stable training with a relatively small target-domain training set.

\vspace{-0mm}\mysection{Experiments}\vspace{-1mm}
We evaluate semantic-segmentation models trained with Attn2mask
on three publicly available datasets: PASCAL VOC 2012 
 (VOC12)\cite{everingham2015pascal}, ImageNet-S \cite{gao2022luss}, and Cityscapes \cite{cordts2016cityscapes}. 
We conduct the main experiments in VOC12, which has been the de facto standard in semantic-segmentation evaluation.
While fully supervised learning in VOC seems close to saturation in terms of accuracy, it is challenging enough to tackle without full annotation and real images.
Additionally, we evaluate Attn2mask's generalizability to 1) a larger-scale segmentation task with ImageNet-S, which was built on ImageNet-1k \cite{deng2009imagenet} and 2) a more task-specific scenario with Cityscapes, a dataset toward urban auto-driving.

\vspace{-3mm}\mysubsection{Dataset generation} \vspace{-2mm}
We used SD 1.4 \cite{sd14} trained on LAION-2B \cite{schuhmann2021laion} without modification except in the domain-adaptation settings.
We generated 512 $\times$ 512-resolution images. 
The attention maps had smaller resolutions but were resized to match the image resolution. SD's U-Net had multiple cross-attention layers and they were averaged after resizing.
We used DDPM \cite{ho2020denoising} sampler with 100 timesteps, and attention maps were taken at the 50th step.
Generation of a sample took around eight seconds with an NVIDIA V100 GPU.
Post-processing was conducted using a Python reimplementation of dCRF \cite{simplecrf} with the default hyperparameters.
\ryoshihab{The whole generation process was done twice with different random seeds to hold out the validation set.}

The prompt set for VOC was created from the image descriptions from COCO captions \cite{chen2015microsoft}.
We curated captions that included VOC class names and their major synonyms that we manually listed up, which made 88,561 captions in total. \ryoshihab{These captions were doubled by prompt augmentation, and finally 80k were selected by CLIPScore \cite{hessel2021clipscore} filtering.}
The prompt set for ImageNet-S was created by a template 
``\texttt{a photo of \$\{class\_name\} with a background}''.
The variable \texttt{\$\{class\_name\}} was randomly chosen from the multiple-defined synonyms per class.
This made the number of possible varieties of the prompts significantly fewer than the needed samples per class (around 100 images/class) for our experiments,
and thus we increased the number of samples by using multiple random seeds per prompt. 
The suffix ``\texttt{with a background}'' helps to reduce the generation of full-foreground images, which is less useful for training.

For Cityscapes, we used a variable-length template 
``\texttt{a photo of \\
\$\{class\_names[0]\}, \$\{class\_names[1]\}, ... , \$\{class\_names[n]\} \\ in cityscape by a dashboard camera}'', where \texttt{class\_names} is a random variable-length list of Cityscapes class names. A number of the class names in Cityscapes were not \kurdyla{everyday} words and thus may not be reflected in the generation, so we replaced ``\texttt{terrain}'' with ``\texttt{sand}'' and ``\texttt{vegetation}'' with ``\texttt{plant}'', and generated 12k samples.
In the adaptation experiments, we used a modified template ``\texttt{a photo of \$\{class\_names[0]\}, \$\{class\_names[1]\}, ... , \\ \$\{class\_names[n]\} in sks cityscape by a dashboard camera}'', where ``\texttt{sks}'' was the reserved word for the domain token [V] in DreamBooth fine-tuning.
More details of this fine-tuning are seen in Appendix.

\begin{table*}[t]
 \caption{Segmentation IoU on VOC12  with the ResNet50 backbone. Our Attn2mask outperformed DiffuMask in the mean.}
 \label{table:pascal}
 \centering
 \vspace{-2mm}
  \begin{tabular}{ccccccccccccccccccccccc}
   \bhline
   Method & bg & aero & bicycle & bird & boat & bottle & bus & car & cat & chair & cow & table \\
   \hline
   \color[rgb]{0.6,0.6,0.6} Real-supervised & \gr 94.4 & \gr 90.0 & \gr 42.4 & \gr 82.1 & \gr 70.5 & \gr 75.8 & \gr 93.4 & \gr 88.1 & \gr 90.7 & \gr 36.5 & \gr 86.5 & \gr 67.2  \\
   \color[rgb]{0,0,0} DiffuMask \cite{wu2023diffumask} & n/a& $\bm{80.7}$ & n/a & $\bm{86.7}$  & 56.9 & n/a & 81.2 & ${74.2}$ & 79.3 & 14.7 & 63.4 & n/a  \\
    Attn2mask (ours)& 88.7 & 65.7 & 34.7 & 82.5 & $\bm{64.7}$ & 62.5 & $\bm{87.0}$ & $\bm{76.0}$ & $\bm{83.2}$ & $\bm{25.0}$ & $\bm{65.3}$ & 49.3 
    \\
   \bhline
   
   \end{tabular}\vspace{2mm}
   \begin{tabular}{ccccccccccccccccccccccc}
   \bhline
   Method & dog & horse & motorbike & person & plant & sheep & sofa & train & tv & \cellcolor[rgb]{0.9,0.9,0.9}mean  \\ \hline
   \color[rgb]{0.6,0.6,0.6}Real-supervised & \gr 86.0 & \gr 90.2 & \gr 87.2 & \gr 85.0 & \gr 68.4 & \gr 88.5 & \gr 57.3 & \gr 84.1 & \gr 78.6 & \cellcolor[rgb]{0.9,0.9,0.9}\color[rgb]{0.3,0.3,0.3} 78.2  \\ 
   \color[rgb]{0,0,0} DiffuMask \cite{wu2023diffumask} & 65.1 & 64.6 & n/a & $\bm{71.0}$ & n/a & 64.7 & $\bm{27.8}$ & n/a & n/a & \cellcolor[rgb]{0.9,0.9,0.9}57.4   \\
    Attn2mask (ours)& $\bm{73.0}$ & $\bm{65.2}$ & 73.6 & 64.8& 35.2 & $\bm{69.0}$ & 13.6 & 62.4 & 58.3 & \cellcolor[rgb]{0.9,0.9,0.9}$\bm{62.2}$  \\
   \bhline 
   \end{tabular}
    \vspace{-4mm}
\end{table*}

\vspace{-3mm}\mysubsection{Configurations of our model} \vspace{-2mm}
\ryoshiha{We implemented two versions of Attn2mask with the same SD but segmentation models with different backbones.
We used DeepLabv3+ \cite{deeplabv3plus2018} with the ResNet50 \cite{he2016deep} and Swin-B \cite{liu2021swin} backbone from the mmsegmentation \cite{mmseg2020} codebase.
For training, we used the BECO codebase \cite{rong2023boundary}.
The reliability threshold coefficient $\alpha$ was set to 1.0, and $\beta$ was set to 0.1.
Finally, we fine-tuned SD using DreamBooth-LoRA \cite{dblora} for transfer to Cityscapes, which we refer to as Attn2mask-LoRA.
We set the base learning rate to 0.01, weight decay to 1e-4, and total batch size to 32 on 2-GPU machines. }

\vspace{-3mm}\mysubsection{Evaluation protocol} \vspace{-2mm}
For PASCAL VOC and Cityscapes, we followed the official protocol for supervised learners, which is applicable without modification for WS learners, including ours.
We used the mean intersection-over-union (mIoU) metric.
For ImageNet-S, we derived a foreground/background (FG/BG) segmentation task
\ryoshiha{in addition to the original 919-class segmentation} by extracting masks that have the same class labels as their whole images.
This was intended to replicate the evaluation protocol of BigDatasetGAN \cite{li2022bigdatasetgan}, in which the authors built their own ImageNet-based segmentation \kurdyla{benchmark,} but it remains unpublished.

\begin{table}[t]
    \hspace{0mm}\begin{minipage}{.6\hsize}
        \caption{Comparisons with the existing diffusion-synthetic semantic segmentation methods on the VOC12 val set. \ryoshihac{$\dagger$: our reproduction.}}
        \vspace{-3mm}
        \label{table:sota}
         \centering
            \begin{tabular}{cccc}
            \bhline 
             Method & Backbone & mIoU \\
             \hline 
            {\it With ordinary-size backbones} & & \\
            DiffuMask \cite{wu2023diffumask} & ResNet50 & 57.4 \\
            Dataset Diffusion \cite{nguyen2023dataset} & ResNet50 & 61.6 \\
             Attn2mask (ours) & ResNet50 & $\bm{62.2}$ \\
            \hdashline 
            {\it With larger backbones} & & \\
              DiffuMask \cite{wu2023diffumask} & Swin-B & 70.6 \\
            Dataset Diffusion \cite{nguyen2023dataset} & ResNet101 &  64.8 \\
            Dataset Diffusion $\dagger$ & Swin-B &  69.5 \\
             Attn2mask (ours) & Swin-B & $\bm{71.0}$ \\
            \bhline 
            \end{tabular}
    \end{minipage}
    \hfill
    \hspace{0mm}\begin{minipage}{.38\hsize}
        \caption{Synthesized dataset quality measured with FID and KID.}
        \label{tab:fid}
         \centering
        \begin{tabular}{cccc}
        \bhline 
             FID$\downarrow$ & Image  & Mask \\
             \hline 
            Ours & $\bm{28.2}$ & $\bm{77.0}$ \\
            Dataset  & \multirow{2}{*}{37.7} & \multirow{2}{*}{ 134.9}  \\
             Diffusion &   & \\
            \bhline 
            \end{tabular}
            \vspace{2mm}\\
            \begin{tabular}{cccc}
            \bhline 
            KID$\downarrow$ & Image  & Mask \\
             \hline 
            Ours & $\bm{0.021}$ & $\bm{0.060}$ \\
            Dataset & \multirow{2}{*}{0.024} & \multirow{2}{*}{0.14} \\
             Diffusion & &\\
            \bhline 
            \end{tabular}
    \end{minipage}  \vspace{-5mm}
\end{table}
\vspace{-2mm}\mysubsection{Results} \vspace{-2mm}
\noindent\hk{\textbf{{Main result on VOC (Tab. \ref{table:pascal} and \ref{table:sota}).}}}
Attn2mask marked mIoU of 62.2\%,  which is, of course, worse than the results by supervised training with the real images but reasonably well for not using real images or manual annotation at all.
It outperformed DiffuMask \cite{wu2023diffumask}, which is \kurdyla{an} important prior work in diffusion-synthetic semantic segmentation without real images with a margin of 4.8\% in the mIoU.
In particular, Attn2mask performs closely to the supervised counterpart in relatively easy classes such as {\it bird} (82.1\% v.s. 82.5\%), {\it bus} (93.4\% v.s. 87.0\%), and {\it cat} (90.7\% v.s. 83.2\%) or contrarily difficult classes where the supervised models do not work well, such as {\it bicycle}  (42.4\% v.s. 34.7\%) or {\it chair}  (36.5\% v.s. 25.0\%). 
However, we also acknowledge that there exists significant degradation of IoU in some classes such as {\it plant} and {\it sofa} compared with the real-image training.
We observed that such classes tend to be generated as parts of backgrounds without being explicitly prompted. This made noise for training these classes, which is the major limitation of our synthetic training.
Table \ref{table:sota} shows additional comparisons using different backbones. 
Attn2mask outperformed all other diffusion-synthetic semantic segmentation training methods without human annotation \cite{wu2023diffumask,nguyen2023dataset}, with ResNet50 and Swin-B settings.
\ryoshihac{Table \ref{tab:fid} shows quality criteria FID \cite{heusel2017gans} and KID \cite{binkowski2018demystifying} of our images and masks w.r.t real VOC, which shows superior label quality of ours to Dataset Diffusion, while the image quality is similar depending on the same SD. }

\begin{table}[t]
    \hspace{0mm}\begin{minipage}{.5\hsize}
        \caption{Comparisons with real-image-based WSSS methods on VOC12 val.}
        \vspace{-4mm}
        \label{table:wsss}
        \centering
        \begin{tabular}{ccccc}
            \bhline 
            & \multicolumn{2}{c}{Training data} &  \\
            Method & Real & Synthetic & mIoU \\
            \hline 
            Pseudo-mask \cite{li2021pseudo} &\checkmark &  & 70.0\\
            ReCAM \cite{chen2022class} & \checkmark & &  70.5\\
            BECO \cite{rong2023boundary} & \checkmark &  & $ 73.7$ \\
            CLIP-ES \cite{lin2023clip} & \checkmark &  & $ 73.8$ \\
            WSSS-SAM \cite{sun2023alternative} & \checkmark &  & $ 77.2$ \\
            MARS \cite{jo2023mars} & \checkmark &  & $\bm{77.7}$ \\
            \hdashline 
            Attn2mask (ours) &  \checkmark  & & $71.7$ \\
            Attn2mask (ours) & & \checkmark & $71.0$ \\
            Attn2mask (ours) & \checkmark & \checkmark & $\bm{75.6}$ \\
            \bhline 
        \end{tabular}
     \vspace{-2mm}
    \end{minipage}
    \hfill
    \hspace{0mm}\begin{minipage}{.48\hsize}
      \caption{\ryoshiha{Ablative analyses of Attn2mask. Evaluated with VOC12.}}
        \vspace{-0mm}
        \label{tab:ablation}
         \centering
         \begin{tabular}{lc}
           \bhline
            & mIoU \\ 
           \hline
           Baseline with ResNet50 & 44.6 \\ 
           + dCRF & 51.2 \\ 
           + Co-training & 58.4 \\ 
           + Adaptive threshold & 58.9\\ 
           + Prompt augmentation & 59.8 \\ 
           + CLIP filtering  & 61.3 \\ 
           + TTA  & 62.2 \\ 
           + Swin-B backbone  & 71.0 \\ 
           \hdashline 
           + Real images  & 75.6 \\ 
           \bhline 
        \end{tabular}
    \vspace{-2mm}
    \end{minipage}\vspace{-1mm}
\end{table}
\noindent\hk{\textbf{Comparisons with real-image-based WSSS {(Tab. \ref{table:wsss}).}}}
\ryoshiha{The WSSS methods listed here used real VOCaug training images.
Our Attn2mask performs closely to existing real-image-based WSSS before 2022 \cite{li2021pseudo,chen2022class} even without using the real images. 
\ryoshihac{While our primary interest is in pure synthetic training, one may be interested in whether the results have room for improvement with the help of real training images. Then, we added the real images from the VOC train set with pseudo-labels created by CAM and IRN \cite{ahn2019weakly} that was distributed along with BECO codebase, and this achieved mIoU 75.6\%, which was competitive in the WSSS setting.}
It is slightly behind of MARS \cite{jo2023mars} and WSSS-SAM \cite{sun2023alternative}, but MARS used a complex object removal strategy and WSSS-SAM used pixel-supervised segment anything model (SAM) \cite{kirillov2023segment} for pseudo-label generation.}

\noindent\hk{\textbf{Ablative analyses {(Tab. \ref{tab:ablation}).}}}
\ryoshiha{
While all of the added modules contributed to the segmentation performance, \ryoshihab{a particularly large} gain was from the robust co-training. We also used test-time augmentation (TTA) and replacement by a larger backbone, which are common techniques in segmentation, for fair comparisons with prior work \cite{wu2023diffumask, nguyen2023dataset}.}

\begin{table}[t]
    \hspace{0mm}\begin{minipage}{.5\hsize}
    \caption{Results in the ImageNet-S FG/BG segmentation evaluated with mIoU. \ryoshihab{Attn2mask enabled annotation-free large-scale segmentation.}}
\vspace{-3mm}
\label{table:imagenets}
 \centering
 \begin{tabular}{cccc}
      \bhline
     & \#Anno- & \#Images &  \\
   Method & tations & 10k & 100k \\
   \hline 
   Real-Supervised & 10k & 82.1 & -- \\
   BigDatasetGAN \cite{li2022bigdatasetgan} & 5k & 74.1 & 74.4 \\
   Attn2mask (ours) & 0 & 63.7 & 66.0 \\
    \bhline 
   \end{tabular}
     \vspace{-2mm}
    \end{minipage}
    \hfill
    \hspace{0mm}\begin{minipage}{.48\hsize}
    \caption{Results in the ImageNet-S 919-class segmentation with the 100k-image settings.}
\vspace{-3.5mm}
\label{table:imagenets919}
 \centering
 \begin{tabular}{cccc}
    \bhline
    & \#Anno- &  \\
   Method & tations & mIoU \\
   \hline 
   Real-Supervised & 10k  &  25.7 \\
   BigDatasetGAN \cite{li2022bigdatasetgan} & 5k & 19.7 \\
   Attn2mask (ours) & 0 & 13.3 \\
   PASS$_s$ \cite{gao2022luss} & 0 & 11.5 \\
    \bhline 
   \end{tabular}
    \vspace{-2mm}
    \end{minipage}\vspace{-5mm}
\end{table}

\noindent\hk{\textbf{{Results on ImageNet-S (Tab. \ref{table:imagenets} and \ref{table:imagenets919}).}}}
Table \ref{table:imagenets} shows the results in ImageNet-S FG/BG segmentation.
We compared Attn2mask with the same network trained with data from  BigDatasetGAN \cite{li2022bigdatasetgan}.
BigDatasetGAN is a semi-supervised generator that relies on manual annotation of generated images to fine-tune GAN for mask generation.
Attn2mask performed competitively without using pixel-label annotations at all, which is \kurdyla{encouraging because shows} the ability to freely increase classes within the range Stable Diffusion can generate. 
\ryoshiha{In ImageNet-S full-class segmentation, a similar tendency was observed despite lower scores with all of the methods due to the hardness of the multi-class segmentation with the extreme numbers of classes, as shown in Table \ref{table:imagenets919}}.

\begin{wraptable}[17]{r}[1mm]{7.3cm}\vspace{-11mm}
\caption{Results in Cityscapes with and without unsupervised adaptation.}
\label{table:city}
 \centering
 \begin{tabular}{cccc}
   \bhline
   Method & Adapt. & Acc. & mIoU \\
   \hline 
   Attn2mask (ours) & & 50.6 & 21.1 \\
   Attn2mask-FT (ours) & \checkmark & 64.1 & 23.8 \\
   Attn2mask-LoRA (ours) & $\checkmark$ & ${75.8}$ & $\bm{25.8}$ \\
    \hdashline 
   MaskCLIP \cite{zhou2022maskclip} && 35.9 & 10.0 \\
   MDC \cite{cho2021picie}  && -- & 7.0\\
   PiCIE \cite{cho2021picie} && -- & 9.7 \\
   ReCo  \cite{shin2022reco} && ${74.6}$ & ${19.3}$ \\
   DiffSeg \cite{tian2023diffuse} && ${76.0}$ & ${21.2}$ \\
    \hdashline 
     MDC \cite{cho2021picie}  &$\checkmark$& 47.9 & 7.1 \\
   PiCIE \cite{cho2021picie}  &$\checkmark$& 40.7 & 12.3 \\
   STEGO \cite{hamilton2022unsupervised} &$\checkmark$& 73.2 & 21.0 \\
    ReCo+ \cite{shin2022reco} &$\checkmark$ & $\bm{83.7}$ & ${24.2}$ \\
    Deep clustering \cite{wang2020deep} &$\checkmark$ & $--$ & ${24.2}$ \\
   \bhline 
   \end{tabular}
\end{wraptable}
\noindent\hk{\textbf{{Domain-adaptative LoRA in Cityscapes (Tab. \ref{table:city}).}}}
LoRA-based adaptation of \\
Attn2mask significantly outperformed the original version and full-finetuning version (Attn2mask-FT), which shows the popular SD's adaptation technique is also useful as an adaptive synthetic training technique.
In contrast to segmenting object images such as VOC, holistic scene understanding in Cityscapes is extremely challenging for WSSS methods using image-level labels. 
Hence, unsupervised segmentation methods \cite{cho2021picie,hamilton2022unsupervised,shin2022reco} are rather intensively developed and we placed reference mIoU values from them.
Attn2mask performs similarly to the existing unsupervised methods even with the significant domain gap between web-crawled training images of SD and the driving images. We \kurdyla{found} the LoRA-based adaptation contributed to the segmentation accuracy, which is encouraging for future development of domain-adaptive diffusion-based segmentation methods.
Attn2mask outperformed the strong unsupervised segmentation methods MaskCLIP \cite{zhou2022maskclip}, PiCIE \cite{cho2021picie} and, STEGO \cite{hamilton2022unsupervised}  in both settings with and without adaptation.
ReCo \cite{shin2022reco} outperformed Attn2mask in the accuracy metric but relied on test-time heavy computation; it used retrieval of relevant images for co-segmentation.

\begin{figure}[t]
    \hspace{0mm}\begin{minipage}{.49\hsize}
      \centering
        \includegraphics[width=.9\linewidth]{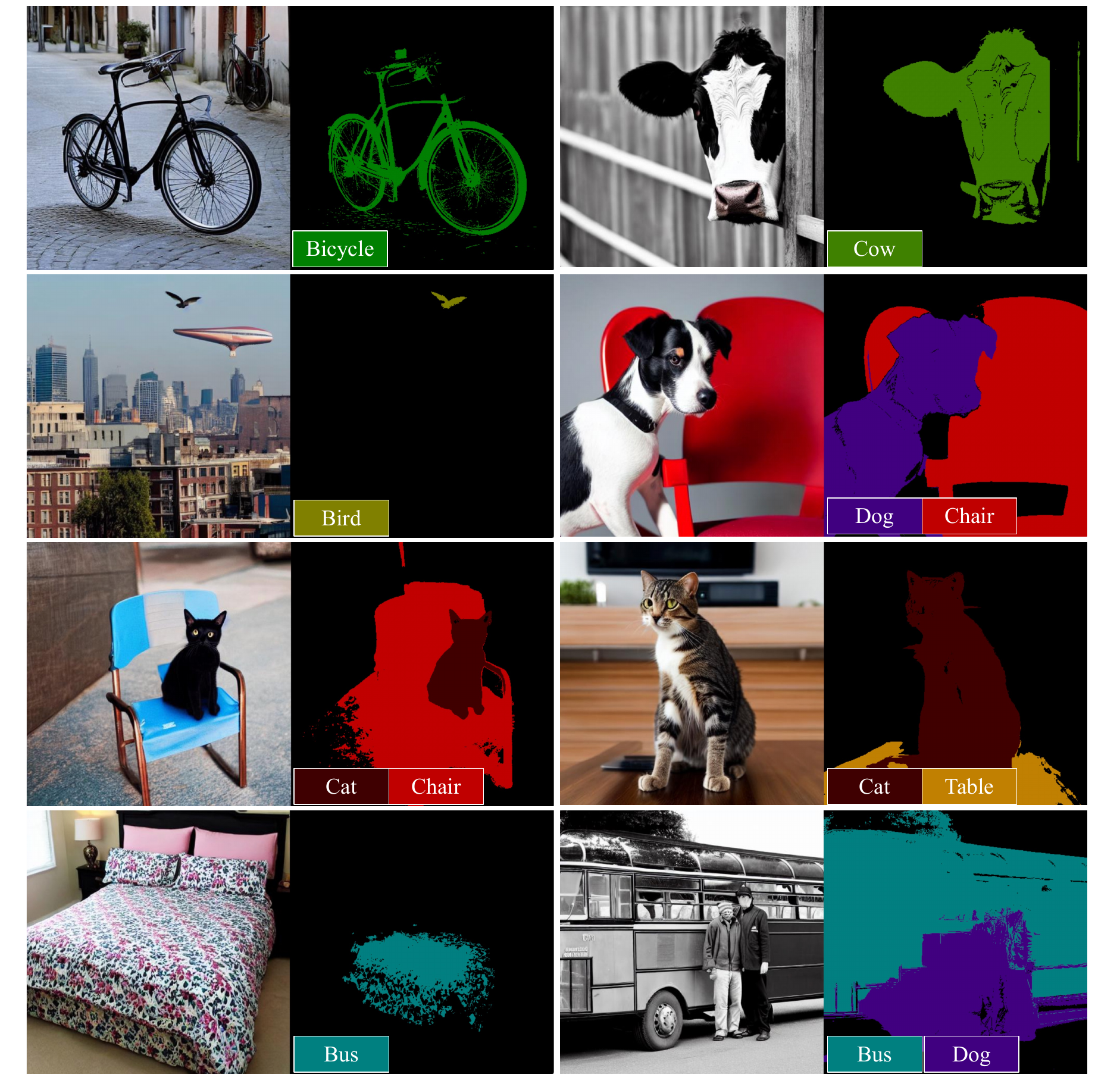}
        \vspace{-4mm}
        \caption{Examples of our training data generated for VOC.}
        \vspace{-3mm}
        \label{fig:dataset}
    \end{minipage}
    \hfill
    \hspace{0mm}\begin{minipage}{.49\hsize}
        \centering
        \includegraphics[width=.98\linewidth]{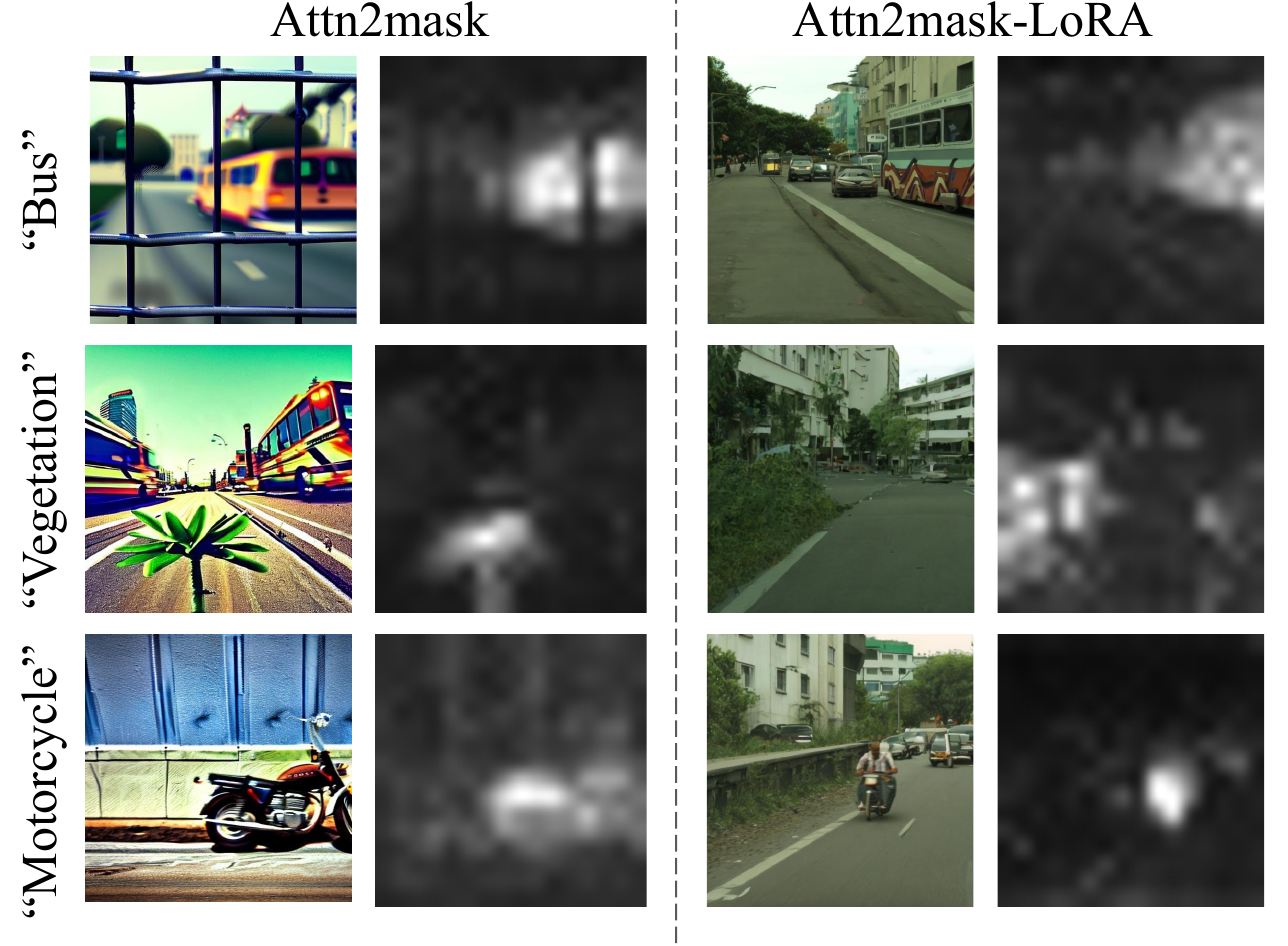}
        \vspace{-2mm}
        \caption{Examples of images and attention maps before and after LoRA-based adaptations in Cityscapes.}
        \vspace{-3mm}
        \label{fig:lora}
    \end{minipage}\vspace{-3mm}
\end{figure}

\vspace{-3mm}\mysubsection{Visualizations} \vspace{-2mm}
Figure \ref{fig:dataset} shows examples of generated VOC training images and labels by \\Attn2mask. In addition to the natural-looking object images, labels are highly accurate except for several missing detailed parts and over-masking of backgrounds where contextual connections exist, e.g., shadows of objects or waves around boats.
We also noticed there exist severe failure cases such as masks not covering whole objects (bottom left) or mismatches of object classes and labels (bottom right).
\ryoshihac{More examples will be seen in Supplementary Material.}

Figure \ref{fig:lora} visualizes synthesized images and attention maps before and after DreamBooth-LoRA-based adaptation.
Before adaptation, image contents and styles were not well-aligned with Cityscapes and the \kurdyla{naturalness} of images was also \kurdyla{somewhat} harmed due to generation in a distant domain from the LAION dataset.
After adaptation, generated images look enough like the images in Cityscapes, while attention maps were kept focused on the objects of interest.
\kurdyla{Interestingly,} the correspondence between attention maps and objects is not corrupted after fine-tuning with the Cityscapes images that are not paired with any captions, which might be because DreamBooth-LoRA reuses the prior knowledge acquired in the original SD well.

\vspace{-2mm}
\mysection{Conclusion and discussion}\vspace{-1mm}
We presented a real-image-and-annotation-free semantic-segmentation method Attn2mask.
Although synthetic training did not outperform real-image-based counterparts to the extent we studied, Attn2mask worked surprisingly well for its purely real-image-free segmentation training, and the gaps between the real and synthetic performances were non-negligibly narrow by incorporating ideas from WSSS literature. 
To summarize our findings, 
1) generative attentions are not always accurate, but we have chances to correct them in the segmentation-training phase by robust co-training (\cref{sec:cot}, \cref{table:pascal} and \ref{table:sota}).
2) Stable Diffusion's ability to synthesize diverse samples can be exploited for segmentation via prompt augmentation or ImageNet-class large-scale generation (\cref{sec:pa}). 
3) In LoRA-based adaptation, Stable Diffusion does not lose its mask-generation ability and thus it is useful for transferring the knowledge to far domains (\cref{sec:da} and \cref{table:city}).
\ryoshihac{The rising SD-synthetic training may cast a question like ``Is SD all we need for WSSS?'' to the community.
We hope that our finding ``WSSS and SD work better together''  lays the groundwork for future research, e.g., WSSS in the era of generative AIs. }
\vspace{-4mm}\ryoshihac{\subsubsection{Potential social impact} Generative models, including SD, can be influenced by social biases in their training data, and our method of training segmentation models with the generated data may inherit the biases. Mitigating such biases is an ongoing research topic, and we have to be careful about deployment in socially sensitive or important usages. We used SD 1.4, which used the LAION dataset that was temporarily retracted for legal and safety concerns \cite{laionsafe}. While Stable Diffusion 1.4 is kept published, we plan to replace it after a Stable Diffusion model trained on safe datasets is released.}

%
%
{
\bibliographystyle{splncs04}
\bibliography{egbib}
}
\end{document}